\documentclass[12pt, letterpaper]{article}
 
\usepackage{amsmath}
\usepackage{amssymb}
\usepackage[longnamesfirst]{natbib}
\usepackage{eqlist} 
\usepackage[final]{graphicx}
\usepackage[dvipsnames]{color}

\usepackage[T1]{fontenc}
\usepackage{lmodern}
\usepackage{textcomp}
\usepackage{microtype}
\usepackage{setspace}

\usepackage{booktabs} 
\usepackage{subfig}
\usepackage{hyperref}
\usepackage{mdwlist}
\usepackage{url}
\usepackage{verbatim}
\usepackage{float}
\usepackage{multirow}
\bibpunct{(}{)}{;}{a}{}{,} 
\DeclareGraphicsExtensions{.pdf, .jpg}

\title{Creating a Real-Time, Reproducible Event Dataset}
\author{John Beieler\thanks{The work in this paper was funded in part by the National Science Foundation under IGERT Grant DGE-1144860, Big Data Social Science. Additional work was completed while at Caerus Associates with
DARPA funding for various portions of the open-source software development.}\\ Human Language Technology Center of Excellence\\ Johns Hopkins University}

\begin{document} 
\maketitle 

\thispagestyle{empty}
\newpage
\thispagestyle{empty}
\begin{abstract}
\begin{normalsize}
\noindent  
The generation of political event data has remained much the same since the mid-1990s, both in terms
of data acquisition and the process of coding text into data. 
Since the 1990s, however, there have been significant improvements in open-source natural language
processing software and in the availability of digitized news content. This paper presents a new, next-generation event
dataset, named Phoenix, that builds from these and other advances. This dataset includes improvements in the underlying news collection
process and event coding software, along with the creation of a general processing pipeline necessary to produce daily-updated data. 
This paper provides a face validity checks by briefly examining the data for the conflict in Syria, and a comparison between Phoenix 
and the Integrated Crisis Early Warning System data.

\end{normalsize}
\end{abstract}

\newpage

\doublespacing
\section{Moving Event Data Forward}

Automated coding of political event data, or the record of who-did-what-to-whom within the context of political actions,
has existed for roughly two decades. The approach has remained largely the same during this time, with the underlying coding
procedures not updating to reflect changes in natural language processing (NLP) technology. These NLP technologies have now
advanced to such a level, and with accompanying open-source software implementations, that their inclusion in the event-data coding
process comes as an obvious advancement. When combined with changes in how news content is obtained, the ability to store and
process large amounts of text, and enhancements based on two decades worth of event-data experience, it becomes clear that political
event data is ready for a next generation dataset. 

In this chapter, I provide the technical details for creating such a next-generation dataset. The technical details lead to a pipeline
for the production of the Phoenix event dataset. The Phoenix dataset is a daily updated, near-real-time political
event dataset. The coding process makes use of open-source NLP software, an abundance of online news content, and other technical
advances made possible by open-source software. This enables a dataset that is transparent and replicable, while providing a more 
accurate coding process than previously possible. Additionally, the dataset's near-real-time nature also enables many applications
that were previously impossible with batch-updated datasets, such as monitoring of ongoing events. Thus, this dataset provides a 
significant improvement over previous event data generation efforts.

In the following sections I briefly outline the history of computer-generated political event data to this point in history. I then
outline what the ``next generation'' of event data should look like. Following this, I discuss the many facets of creating a real-time political event dataset, mainly from a technological and infrastructure standpoint. Finally, the paper concludes with a brief empirical view of the Phoenix event dataset, which is the output of the previously-discussed technological pipeline.

\section{The History of Event Data}

Political event data has existed in various forms since the 1970s. Two of the most common political event datasets were the World Event
Interaction Survey (WEIS) and the Conflict and Peace Data Bank (COPDAB) \citep{azar, weis}. These two datasets were eventually
replaced by the projects created by Philip Schrodt and various collaborators. In general, these projects were marked by the use of the Conflict
and Mediation Event Observations (CAMEO) coding ontology and automated, machine-coding rather than human coding \citep{cameo, cameo2}.
The CAMEO ontology is made up of 20 ``top-level'' categories that encompass actions such as ``Make Statement'' or ``Protest'', and contains over 200 total
event classifications. This ontology has served as the basis for most of the modern event datasets such as the Integrated Crisis Early Warning System (ICEWS) \citep{obrien}, 
the Global Database of Events, Language, and Tone (GDELT)\footnote{\href{gdeltproject.org}{gdeltproject.org}}, and the Phoenix dataset presented in this paper.

This type of data can prove highly useful for many types of studies. Since this type of data is inherently atomic, each observation is a record of a single event between
a source and a target, it provides a disaggregated view of political events. This means that the data can be used to examine interactions below the usual monthly or yearly
levels of aggregation. This approach can be used in a manner consistent with traditional hypothesis testing that is the norm in political science \citep{gleditsch1, gleditsch2, goldstein2}. Additionally, event
data has proven useful in forecasting models of conflict since the finer time resolution allows analysts to gain better leverage over the prediction problem than is possible
when using more highly aggregated data \citep{mine, forecasting1, forecasting2, forecasting3}. Finally, the advent of daily-updated event data has led to many novel uses such as watchboarding or dashboarding. The goal in
these situations is to provide an easy to understand interface that analysts can use to quickly monitor ongoing or emerging situations around the world. These applications provide a new
frontier for event data that has not been considered much until this point. 

The status quo of TABARI-generated, CAMEO-coded event data, which was established in the early 2000s, has remained with little change.\cite{schrodt2012} outlined many potential advances in the generation
of political event data. These advances are things such as realtime processing of news stories, the incorporation of open-source natural language processing (NLP)
software, and enhancements in the automated coding structure. Two publicly-available datasets, GDELT and ICEWS, have each attempted
to implement some, or all, of these changes in their respective data-generating pipelines. In terms of goals, the ICEWS project seems
closest to sharing the vision of the Phoenix dataset. A more
in-depth comparison of Phoenix and ICEWS is presented in a later
section.
In short, the goal of the project presented in this chapter is to implement most of the improvements suggested in \cite{schrodt2012}. 

\section{Event Data: The Next Generation}

One of the defining traits of previous event-data projects is the method through which they were generated. 
The original datasets such as WEIS and COPDAB were created by human coders who read news stories and coded
events. Future datasets such as KEDS and Phil Schrodt's Levant dataset were created using automated coding
software, such as KEDS or TABARI, and news stories download from content aggregators such as Lexis Nexis or
Factiva. Both pieces of coding software made use of a technique referred to as shallow parsing \citep{sparse}. Shallow
parsing is best understood in contrast to a deep parsing method. In deep parsing, the entire syntactic structure
of a sentence is used and understood. This syntactic structure includes things such as prepositional phrases, direct
and indirect objects, and other grammatical structures. A shallow parse, however, focuses solely on, as the name implies,
shallow aspects such as the part of speech of the words within the sentence. 

The second major dimension that differentiates event datasets is how news content was acquired. For WEIS and COPDAB this was
as simple as subscribing to the New York Times and coding from there. Later datasets, such as those created in conjunction with the
Kansas Event Data Project, obtained historical content from aggregators, as mentioned above. This difficulty of this process 
changed at various points in time, with something like full automation possible at some points while human downloading of stories
was required at others. There are often gaps in this historical content since the content aggregators catalog of different news
services changes at various points and is often fairly limited. Updating datasets based on this type of content was also fairly
labor intensive since new content had to be downloaded, cleaned, and run for every update. While orders of magnitude faster than
human coding, this remained an involved process.

Taken together, these two aspects of event data generation, shallow parsing and content acquisition, form the basis for where
the next generation of political event data can improve upon previous efforts. In short, a shift to deep parsing based on relatively
recent advances in open-source natural language processing software, combined with realtime acquisition of news content and 
aggressive strategies for acquiring historical material, provide the motivation for the next generation of political event data. The
following section provides greater detail regarding the implementation of these new features. 

\section{Building A Pipeline}

The following sections outline the multiple aspects that go into building a near-real-time pipeline for the creation of political event data.
First, I provide a discussion of the considerations that went into the architecture of the software used to create the data. Next, I outline the various
advances that have been made in the data collection and processing steps. Finally, a discussion of the challenges and obstacles faced when deploying such
a software pipeline is presented.

\subsection{Considerations}

There are three main considerations at play when designing software surrounding the Phoenix event data pipeline: modularity, composability, and 
reproducibility. In short, no one part of the pipeline should be hardcoded to operate within the pipeline, implying other pieces are easily replaced
by new and/or better alternative, and the pieces should operate in such a manner that reproducing the exact steps used to create the final dataset
is transparent and understandable to those within the broader event data community. Towards this end, the pieces of software are \emph{modular} in 
nature; each piece can stand on its own without relying on another other piece of software in the stack. These modular pieces lead to a system that
is \emph{composable}. As pieces can stand on their own, parts of the system can be replaced without affecting the rest of the system in an major way. 
Finally, the modular and composable nature of the pipeline leads to a system that is inherently \emph{reproducible}. In many ways, the code itself serves
as documentation for reproduction. If the versions of the various pieces are noted, all that is necessary to reproduce the pipeline is to link the 
correct versions of each module together. Proper design nearly guarantees reproducibility of the data generating process. 

\subsection{Advances}

\subsubsection{PETRARCH}

PETRARCH (Python Engine for Text Resolution And Related Coding Hierarchy) is the new generation of event-data coding software that is
the successor to the TABARI software. As noted in the previous sections, the major advance of this next generation of event data coding
is the incorporation of a ``deep parse'' that enables more advanced analysis of the syntactic structure of sentences. In PETRARCH's case,
this deep parse is provided by the Stanford NLP group's CoreNLP software \citep{stanford}. CoreNLP provides information regarding part-of-speech tags for
individual words, noun and verb phrase chunking, and syntactic information regarding the relation of noun and verb phrases. Figure 1 provides
an example of what information CoreNLP outputs, while Figure 2 provides an example of the input that PETRARCH accepts.  

\begin{center}
\begin{figure}[H]
\includegraphics[height=7in, width=6in]{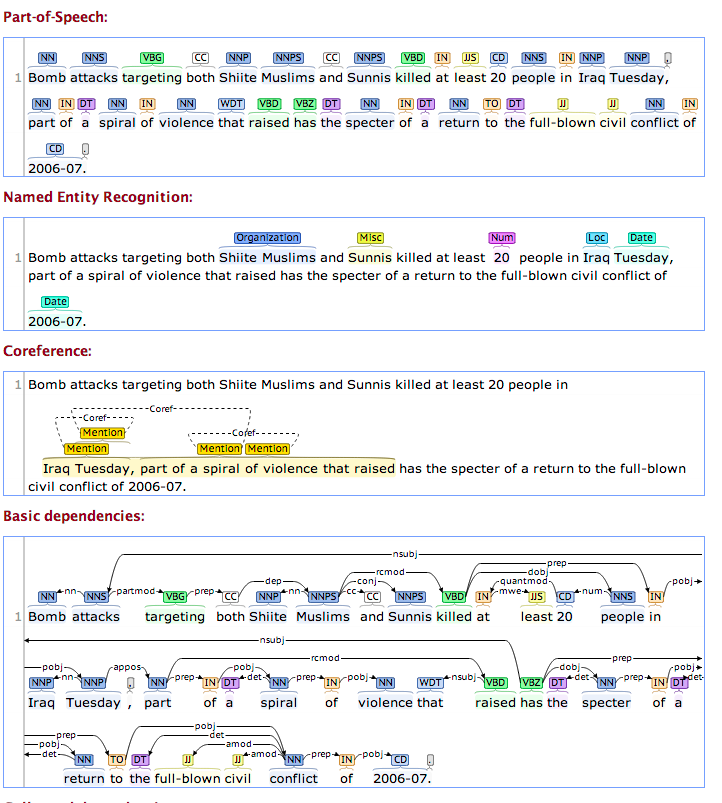}
\caption{CoreNLP Annotations}
\end{figure}
\end{center}

\begin{center}
\begin{figure}[H]
\includegraphics{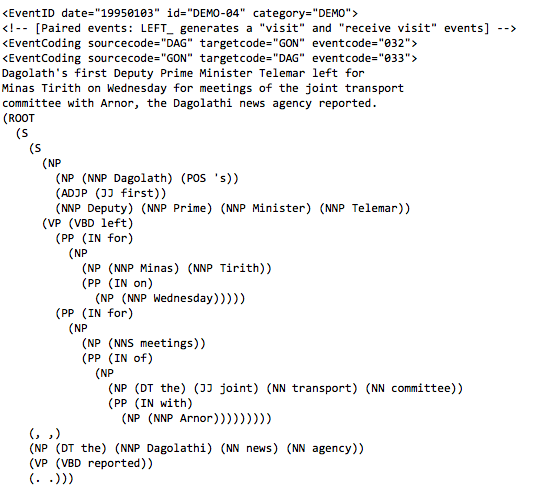}
\caption{Parse Tree Input for PETRARCH}
\end{figure}
\end{center}

The main advantage that this deep parse information provides for the current iteration of PETRARCH is improved noun and verb phrase disambiguation.
At its heart, PETRARCH is still software to perform a lookup of terms in a set of text dictionaries. Given this, if the terms identified by the program
are incorrect then the final event coding will also be incorrect. The list of noun and verb phrases created as output by CoreNLP increases the probability
that the terms used to match in the dictionaries are ``good.'' Thus, in theory, PETRARCH coding should be more accurate due to a more accurate identification
of noun phrases, which translates to actor lookups, and verb phrases, which translates to event code lookups. To put it bluntly, PETRARCH operates in much the
same manner as TABARI, but offloads the issues of dealing with syntactic and grammatical structures to purpose-built software. 

The downside of the use of deep parse information is the increased computational load required to process the news stories. TABARI is capable of processing upwards
of 10,000 sentences per second, whereas CoreNLP can process less than 100 sentences per second\footnote{This is a relatively recent speedup in the CoreNLP software. The
CoreNLP parse was previously the substantial bottleneck in processing. With the release of version 3.4, however, the Stanford NLP team introduced a new shift-reduce parser
that sped up the processing time by a substantial amount.} and PETRARCH codes around 100 sentences per second. The slowness of CoreNLP is due to the complexity of applying 
the parsing models to English-language sentences. PETRARCH is slow for a variety of reasons, foremost among them being the use of the Python programming language as opposed to
the C++ language TABARI uses.\footnote{Profiling and modifications of PETRARCH in order to enhance performance are ongoing at the time of writing. The most likely way forward
for PETRARCH is to rework important parts of the codebase to make use of the C programming language, which can be much faster than the higher-level Python code.} Additionally, speed issues likely arise from the internal data representation of PETRARCH as opposed to TABARI; with TABARI making 
use of more efficient lookup algorithms and data structures.

\subsubsection{PETRARCH2}

PETRARCH2 represents a further iteration upon the basic principles seen in PETRARCH, mainly a deep reliance on information from a syntactic parse tree. 
The exact operational details of PETRARCH2 are beyond the scope of this chapter, with a complete explanation of the algorithm available in \cite{petr2}, 
it should suffice to say that this second version of PETRARCH makes extensive use of the actual structure of the parse tree to determine source-action-target event codings. In other words, PETRARCH still mainly focused on parsing noun and verb phrase chunks without fully integrating syntactic 
information. In PETRARCH2 the tree structure of sentences is inherent to the coding algorithm. Changing the algorithm to depend more heavily on the tree 
structure of the sentence allows for a clearer identification of actors and the assignment of role codes to the actors, and a more accurate 
identification of the who and whom portions of the who-did-what-to-whom equation. The second major change between PETRARCH and PETRARCH2 is
the internal category coding logic within PETRARCH2. In short,
PETRARCH2 allows for interactions of verbs to create a different
category classification than either verb on its own would produce.
For PETRARCH, such things would have to be defined explicitly within
the dictionaries. In PETRARCH2, however, there is a coding scheme
that allows verbs like ``intend'' and ``aid'' to interact in order
to create a different coding than either verb on its own would create.\footnote{This explicit workings of this can be viewed at \url{https://github.com/openeventdata/petrarch2/blob/1.0.0/petrarch2/utilities.py\#L265}.} Additionally, PETRARCH2 brought about a refactoring and 
speedup of the code base and a reformatting of the underlying verb dictionaries. This reformatting of the dictionaries also included
a ``cleaning up'' of various verb patterns within the dictionaries.
This was largely due to changes internal to the coding engine such
as the tight coupling to the constituency parse tree and the verb
interactions mentioned above.
This change in the event coder software further demonstrates the modular 
and composable nature of the processing pipeline; the rest of the processing architecture is able to remain the same even with a relatively major shift
in the event coding software.

\subsubsection{Realtime News Scraping}

There are several ways that the scraping of news content from the web can occur. A system can sit on top of an aggregator such
as Google News, use a true spidering system that follows links from a seed list, or can pull from a designated
list of trusted resources. Each system has its benefits and challenges. The use of an aggregator means that a project is subject
to another layer of complexity that is out of the user's control; those making use of Google News have no say over how, and what,
content is aggregated. Implementing a full-scale web spider to obtain news content is a labor and maintenance intensive process that
calls for a dedicated team of software engineers. This type of undertaking is beyond the scope of the current event data projects. 
The final option is to use a list of predefined resources, in this case RSS feeds of news websites, and pull content from these 
resources. For the purposes of the realtime event data discussed herein, I have settled on the final option. 

The conceptual implementation of a web scraper built on top of RSS is relatively simple. Given a defined list of RSS feeds, pull 
those feeds at a fixed time interval and obtain the links to news stories contained within the feeds. The final step is to then
follow the links to the news stories and obtain the news content. The
relevant content is obtained through the use of the Python library \texttt{Goose}.\footnote{\url{https://github.com/grangier/python-goose}} \texttt{Goose} works through a series of heuristic rules to identify which portions of the web page contain content rather than things such as navigation links and advertisements. These heuristics operate on the HTML
tags within a page, and the inherent tree-structure of the relationships
between these tags. I, with the contributions of others, created an open-source software 
implementation of this RSS scraping concept which works well for a couple hundred RSS feeds.\footnote{\href{https://github.com/openeventdata/scraper}{https://github.com/openeventdata/scraper}}
As the scope and ambition of the
event data project grew, however, it became clear that this implementation is less than adequate for the task. Thus, the final 
scraper product, named atlas, moved to a distributed worker queue model that continuously queries RSS feeds to check for new links and
consumes new content as it becomes available.\footnote{\href{https://github.com/johnb30/atlas}{https://github.com/johnb30/atlas}} This architecture has enabled the scraping of over 500 RSS feeds in both English and Arabic.
This distributed architecture also allows for nearly infinite scalability; workers can move from process on an individual server
to process on a cluster of servers. 

This scraped content is stored in a NoSQL database, specifically a MongoDB instance, due to the inherently flexible nature of
NoSQL databases. The lack of a predefined schema allows requirements and storage strategies to change and update as the scraping
process matures and more knowledge is gained. This is especially important given the ever changing nature of web scraping. Some
sites can move from being viable sources of information to no longer being useful or relevant. Sometimes sites update and break the
scraping process. A flexible storage format allows for this information to be accommodated as it arises. 

\subsubsection{Geolocation}

The final additional piece of information necessary for a modern event dataset is the geolocation of the coded events. The geolocation
of event data is difficult from both a technological and ontological perspective. First, from an ontological standpoint, deciding which
location to pick as \emph{the} location for an event is often difficult. For example, a sentence such as ``Speaking from the Rose Garden,
President Obama denounced the Russian actions in Syria'' provides several possible locations: the Rose Garden, Syria, and even, possibly, Russia.
It is also possible for an event to have no location. This problem relates to the ``aboutness'' of an article.
In the above example, the statement event of President Obama denouncing Russia should
likely be coded as \emph{not} having a location. The second difficulty is the technological issues at play when geolocating place mentions.
First, geolocation must sit on top of named entity recognition, which is itself a fragile process. Once these location identities are 
identified, they must be resolved to their latitude and longitude coordinates. These lookups are difficult since any process must disambiguate
between Paris, Texas and Paris, France or between Washington state and Washington D.C. Finally, event data coding currently works at the sentence level, 
which restricts how much information can be discerned when using the entirety of an article's text. 

In order to achieve geolocation, the Phoenix pipeline currently makes use of the CLIFF\footnote{\url{http://cliff.mediameter.org/}{http://cliff.mediameter.org/}} software, which itself sits on top of the CLAVIN\footnote{\url{https://clavin.bericotechnologies.com/}{https://clavin.bericotechnologies.com/}} software. These programs use heuristics to disambiguate place name mentions and aid in choosing the
specific place that an article is about, thus aiding in solves the ``aboutness'' problem. The process is not perfect however, so
the accurate geolocation of event data is still very much an open problem.

\subsubsection{An Integrated Pipeline}

To make all the various pieces communicate, a comprehensive pipeline is necessary in order to successfully coordinate the various
tasks. Specifically, there are three main pieces of software/technology that must communicate with each other: PETRARCH,
Stanford's CoreNLP software, and the MongoDB instance. For the realtime data component, the web scraper must also fit into this
system. The overall flow of this pipeline is demonstrated in the figure below.

\begin{center}
\begin{figure}[H]
\includegraphics[height=5in, width=6in]{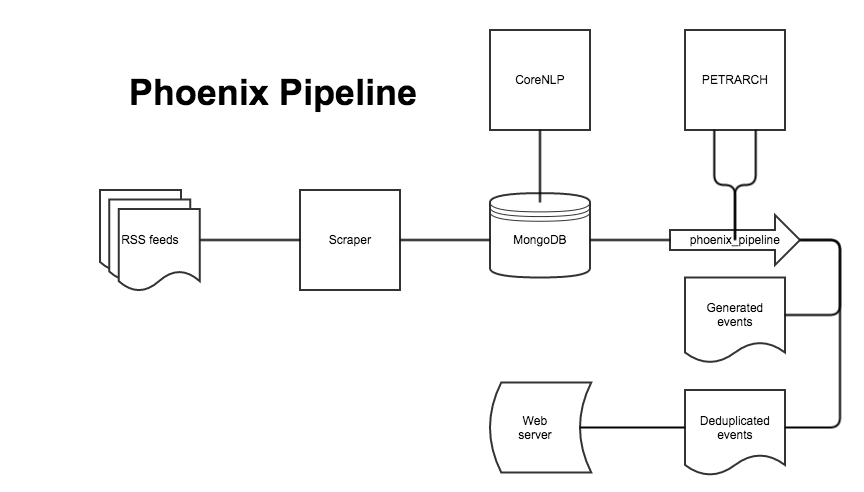}
\caption{Phoenix Pipeline Flow Diagram}
\end{figure}
\end{center}

The modular nature of this pipeline allows for various pieces to be run independently of each other. For instance, content can
be obtained and processed through CoreNLP with the resulting parse stored in a database. This derived parse information can then
be fed into PETRARCH several times following updates to the underlying dictionaries or to the software itself. Likewise, if the
scraping software needs an update or a move to a different architecture, as happened with this project, the rest of the pipeline
can carry on as normal since the other pieces are agnostic to how a single aspect of the pipeline functions.

\subsection{Challenges and Obstacles}

While the features discussed in the previous section provide a significant change from previous generations of event data, moving the
field forward also comes with some unexpected challenges. Issues of processing time, data/software versioning, updating dictionaries, and 
reconceptualizing how event data is coded come into play when moving the event data research program forward. Each of these issues is a
difficult problem when taken alone, when combined the obstacles can seem unsurmountable. Future iterations of event data will need to
consider and address each of these issues.

One of the biggest unforeseen issues when moving from a shallow to a deep parse was the exponential increase in processing time. The TABARI
program was extremely fast for two reasons: it is highly optimized C++ code and the shallow parsing markup is a speedy operation. PETRARCH
requires a deep parse generated by software such as CoreNLP. CoreNLP takes a large amount of time to complete a deep parse of news
stories.\footnote{The processing time of CoreNLP has improved with recent versions of the parsing software.} This means that the processing load
for realtime data updating is more than a single consumer computer can handle. It also means that processing large amounts of historical text
takes a significant amount of time.\footnote{Processing 3.6 million news stories took roughly four days to complete.}

Processing realtime data also means that the relevant actors are often changing. For example, during the development process of this event data project
the Islamic State of Iraq and the Levant (ISIL) became a major actor in the Middle East. ISIL and its leadership were not encoded in the actor
dictionaries used in any event data project. Updates to the dictionaries to include these actors lead to a near doubling of events coded in the relevant
countries. This presents a serious issue for the sustainability of realtime coding; dictionary updating is a labor intensive process that lacks
much of the appeal to potential funders that other projects have. Automated entity extraction is an area of active research that can help in this
situation\footnote{See \cite{getoor} for an example of entity resolution.}, but the main step, actually creating new actor 
codes for the relevant entities, is one that currently still needs a ``human in the 
loop.''

The constantly changing nature of the constituent parts of the event data coding process (both software and text dictionaries) creates a problem
for various parties interested in using event data. A balance must be struck between moving quickly to satisfy users more interested in the realtime
updates, while preserving data stability for those users that need a long time series of data. One approach, which has been embraced by the Open
Event Data Alliance, is to aggressively version every product, be it software or text, that is produced and relates to the event data coding process.
This means that data can be coded using a mix-and-match approach and the version numbers of the various parts can be indicated in documentation. This
also allows for a differentiation between ``bleeding-edge'' versions of the data and stable/maintenance 
releases.\footnote{This versioning scheme pulls heavily from the software development world where products will often have a ``nightly'' build that 
incorporates all of the newest changes and a ``stable'' release that is meant for broad, public use.}

Finally, moving into realtime event coding raises issues of whether the traditional who-did-what-to-whom format is still the best data structure
for further development. Pulling news content from the web increases both the amount and diversity of information obtained. Much of this material
contains sentences that are code-able by the PETRARCH software but that don't produce events in the standard who-did-what-to-whom format. For example,
some events such as protests or statements might not have an explicit target actor. This differs from previous event data which focused mainly on dyadic
interactions between state actors. In addition to the actor issues, the new source material raises questions regarding what type of actions should be coded.
Utilizing existing coding ontologies such as CAMEO restricts the code-able actions to a relatively small subset of all political interactions. 

\section{Production-Ready Versions}

The pipeline described above is a relatively complicated software system; the various features described such as modularity lead to a disconnected system 
that requires knowing a large amount of detail about a high number of components. To help ease this burden, I have created, or participated in the 
creation, of open-source software tools to help with the deployment of the various components of the pipeline.

\subsection{EL:DIABLO}

\begin{center}
\begin{figure}[H]
\includegraphics[height=3in, width=6in]{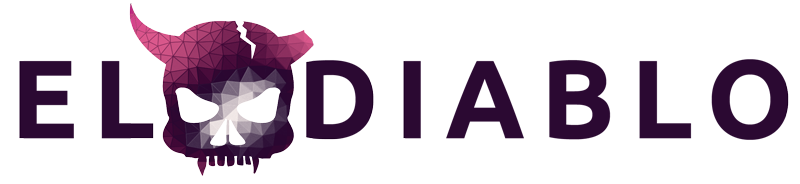}
\caption{EL:DIABLO}
\end{figure}
\end{center}

EL:DIABLO\footnote{\url{https://github.com/openeventdata/eldiablo}} is, at its heart, a script to setup a
virtual machine with each of the software components configured, installed, and linked properly. This virtual machine is a ``computer within a computer''
that allows users to exactly replicate the production pipeline used to create the daily-updated Phoenix data. This virtual machine servers two purposes. First, it allows the fulfillment of each of the main goals described in section 4.1; the components are able 
to stay modular with the entire pipeline being reproducible without each user having to maintain a knowledge of how the entire pipeline functions. 
Second, the script used to create the virtual machine servers as documentation and an example for how one would deploy the pipeline in a setting outside 
a virtual machine.

\subsection{hypnos}

For many applications, deploying the entire pipeline as done via EL:DIABLO is drastic overkill. For instance, a user might want to process a set of 
existing texts or may wish to insert the main event data coding software, PETRARCH or PETRARCH2, into an existing infrastructure. To aid in this, 
\texttt{hypnos}\footnote{\url{https://github.com/caerusassociates/hypnos}} was created to deploy the two
minimal components necessary to code event data: the event coder and CoreNLP. These two components are wrapped in a REST API, which allows users to 
make HTTP requests. The components are wrapped as Docker\footnote{\url{https://www.docker.com/}} containers, which allows
for easy deployment and transportability of applications. Thus, with a single command users are capable of standing up an easy-to-use API around the two
main event coding components.

\section{The Phoenix Dataset}

The Phoenix dataset is an attempt to take both the new advances in event data described above, along with decades of knowledge regarding best
practices, in order to create a new iteration of event data. The dataset makes use of 450 English-language news sites, which are each scraped
every hour for new content. New data is generated on a daily basis, coded according to the CAMEO event ontology, with an average of 2,200 events generated per day. The full dataset examined 
here contains 254,060 total events spread across 102 days of generated data. Based on publicly available information, the project also makes use of the most up-to-date actor dictionaries of 
any available machine-coded event dataset.\footnote{All dictionaries are available at \url{https://github.com/openeventdata/Dictionaries}.} 

The dataset currently contains 27 columns: ``EventID'', ``Date'', ``Year'', ``Month'', ``Day'', ``SourceActorFull'', ``SourceActorEntity'', 
``SourceActorRole'', ``SourceActorAttribute'', ``TargetActorFull'', ``TargetActorEntity'', ``TargetActorRole'', ``TargetActorAttribute'', 
``EventCode'', ``EventRootCode'',  ``QuadClass'', ``GoldsteinScore'', ``Issues'', ``ActionLat'', ``ActionLong'', ``LocationName'', 
``GeoCountryName'', ``GeoStateName'', ``SentenceID'', ``URLs'', ``NewsSources.'' While there are columns included for geolocation of events,
this feature is not fully implemented due to the difficult nature of accurately geolocating event data.

The \texttt{*ActorFull} columns include the full actor coding, which is made up of several three-letter CAMEO codes strung together. 
\texttt{*ActorEntity} breaks out the top-level code, which is usually a country code but can also be ``IMG'' for international militarized group, 
``IGO'' for inter(national)  governmental organizations, or ``MNC'' for multinational corporations. \texttt{*ActorRole} includes codes like 
``GOV'', ``MED'', ``EDU'',  ``MIL'', and \texttt{*ActorAttribute} includes modifiers, such as ``MOS'', ``INS'', ``ELI'' (Muslim, insurgent, and 
elite).

\texttt{EventCode} is the full CAMEO code, while \texttt{EventRootCode} is the 20 top-level CAMEO categories. The \texttt{QuadClass} is an 
updated version of the quad class divisions seen in other event datasets. The changes include the creation of a 0 code for CAMEO category 01 
(``Make a Statement''), rather than counting 01 as verbal cooperation, as well as several lower-level codes changing quad classes. Previous 
quadclass implementations sliced the CAMEO categories in a linear fashion. This new implementation takes into consideration what the CAMEO 
categories actually suggest in terms of material or verbal conflict/cooperation. In this scheme, 0 is ``Neutral,'' 1 is ``Verbal Cooperation,'' 2 is ``Material Cooperation,'' 3 is ``Verbal Conflict,'' and 4 is ``Material Conflict.'' The categories are as follows:

\singlespacing
￼￼\begin{center}
\begin{table}[H]
\caption{CAMEO-to-Quad Class Conversions}
\label{tab:results}
\centering
\begin{tabular}{c c c} 
\hline\hline           
     Root CAMEO  &    Description  & Quad Class      \\
\hline
01 &  Make Public Statement & 0  \\
02 &  Appeal    &    0  \\
03 &  Express Intent to Coop   &  1  \\
04 &  Consult    &    1  \\
05 &  Engage in Dip Coop   & 1  \\
06 &  Engage in Material Coop  & 2  \\
07 &  Provide Aid    &    2  \\
08 &  Yield    &    2  \\
09 &  Investigate    &    3  \\
10 &  Demand    &    3  \\
11 &  Disapprove    &    3  \\
12 &  Reject    &    3  \\
13 &  Threaten    &    3  \\
14 &  Protest    &    4  \\
15 &  Exhibit Force Posture  & 4  \\
16 &  Reduce Relations    &    3  \\
17 &  Coerce    &    4  \\
18 &  Assault    &    4  \\
19 &  Fight    &    4  \\
20 &  Use Unconventional Mass Violence    &    4  \\
\hline\hline
\end{tabular}\label{tab:cameo}
\end{table}
\end{center}

\doublespacing
\noindent
The \texttt{GoldsteinScore} variable is the same, standard scale used in previous datasets \citep{goldstein}.\footnote{As a note, the ``Goldstein scale'' used for CAMEO-coded data is not the same as the scaled presented in \cite{goldstein}, which was designed for the WEIS coding ontology. Instead it is a modification made by a graduate student from Kansas University in the early 2000s.}
The final column relating to event
actions is codes for \texttt{Issues}. These issues are based on simple keyword lookups and serve as a mechanism to add further context to
a CAMEO code. For instance, a statement (CAMEO code \texttt{01}) might be about a specific topic such as education. 

The final three columns include citation information for the events, including which news sources reported the event, the URLs for the story, an 
internal database ID for the stories, and which sentence in each story contained the coded event.

\subsection{Description}

In order to obtain a broad picture of how the data is structured over time, Figure 1 presents a time series of daily counts of events
within the Phoenix dataset. There are three main interesting aspects presented in this figure. First, the number of events generated
stays relatively stable over time. Second, there is some apparent weekly periodicity in the data with lower numbers generated on the
weekends. Finally, there are points where the number of events generated drops to near zero. This is the result of either server
failures or software bugs in the web scraper and is a peril of maintaining realtime software.

\begin{center}
\begin{figure}[H]
\includegraphics{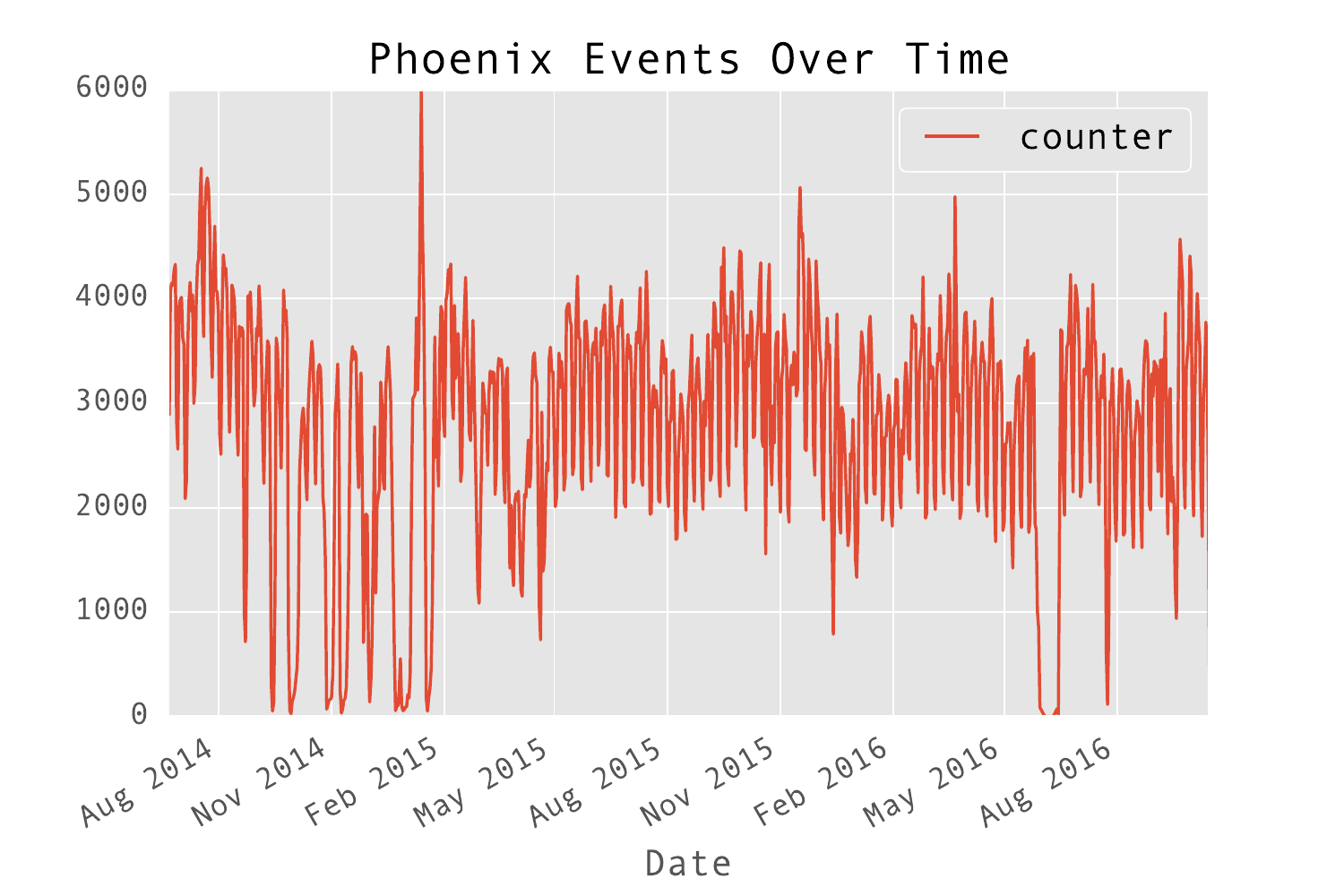}
\caption{Total Phoenix Events Over Time}
\end{figure}
\end{center}

Another piece of useful context is what sources are generating a large portion of the events. Figure 2 shows this information. World News network of sites\footnote{\href{http://wn.com/}{http://wn.com/}} generates the most events, roughly a third. This
is likely due to continuous updates and content that is relevant and code-able under the CAMEO ontology. The other top sources
are made up of sites such as Today's Zaman\footnote{\href{http://www.todayszaman.com/home}{http://www.todayszaman.com/home}} along with sites one would 
expect such as Xinhua and Google News. 

\begin{center}
\begin{figure}[H]
\includegraphics[width=6in]{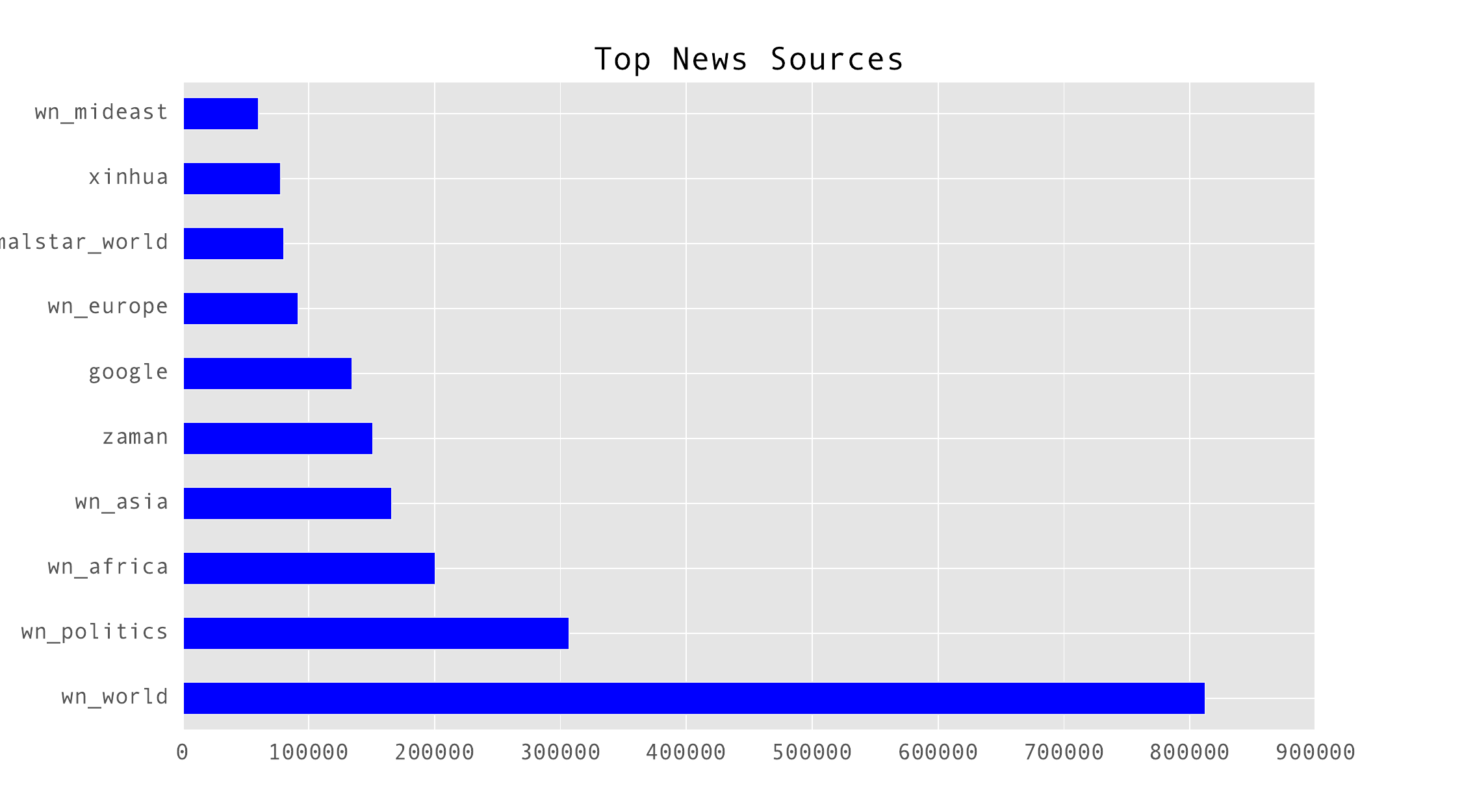}
\caption{Top 10 Sources of Events in Phoenix}
\end{figure}
\end{center}

\subsection{Events}

As has been noted, events are coded on two primary dimensions: event codes and actors. Most political event datasets are dominated
by low-level political events that lack a strong valence. These are usually routine events such as statements that occur often. 
Figures 4 and 5 show the breakdown of event types within the current Phoenix data, both of which confirm this existing pattern.
The addition of the \texttt{0} quad class category was designed to capture these types of events so that they can be easily removed
to allow end users to easily focus on more substantive political events. Following these lower-level event types, the event codes
19 and 17, ``Fight'' and ``Coerce'' respectively, are the next most common. The prevalence of 19 codes is unsurprising given that
the underlying dictionaries were structured in such a way that many events defaulted to this category.\footnote{This is slowly 
changing as can be seen by the work documented at \href{https://github.com/openeventdata/Dictionaries/pull/9}{https://github.com/openeventdata/Dictionaries/pull/9}.}

\begin{center}
\begin{figure}[H]
\includegraphics{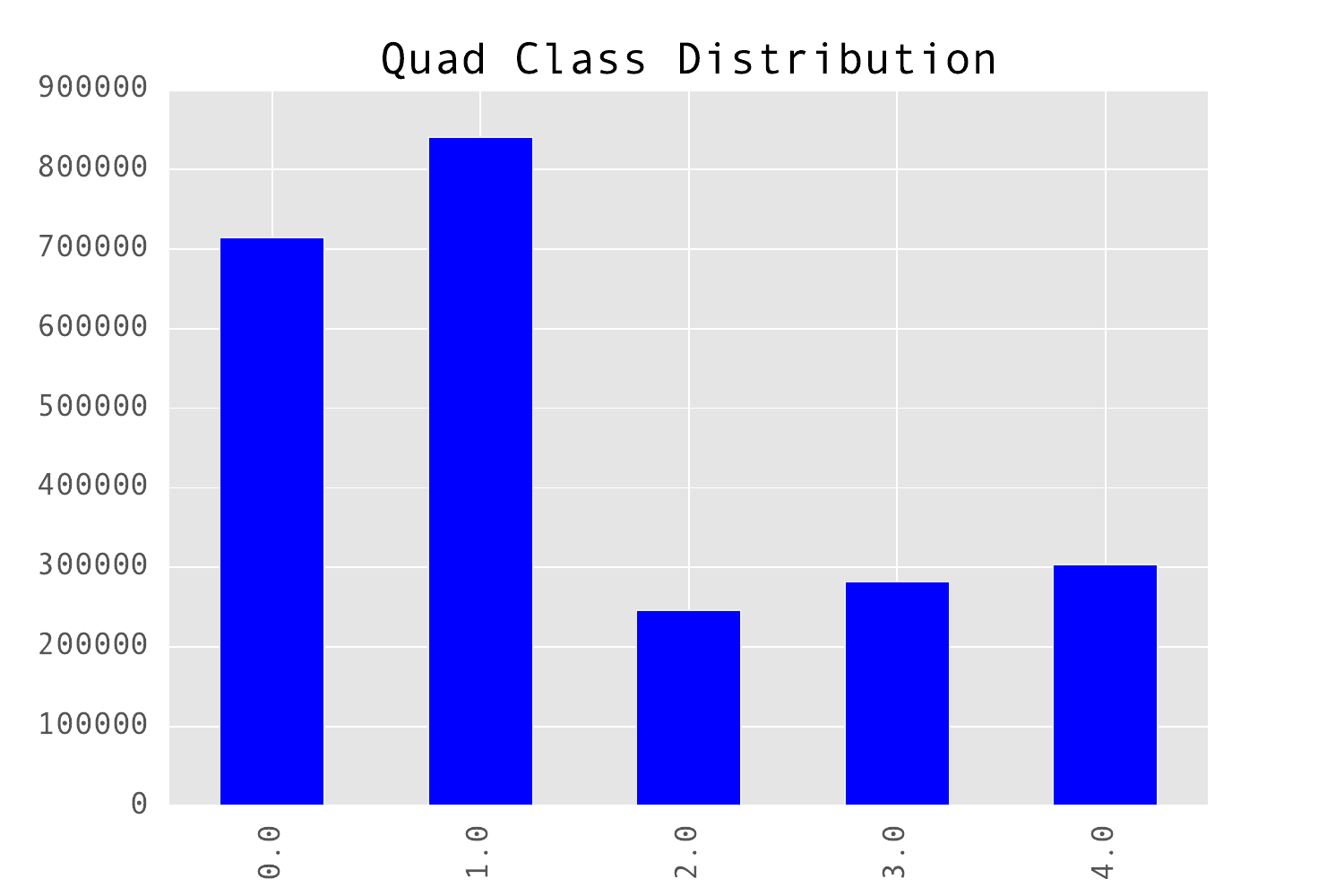}
\caption{Distribution of Quad Class Values in Phoenix}
\end{figure}
\end{center}

\begin{center}
\begin{figure}[H]
\includegraphics[width=6in]{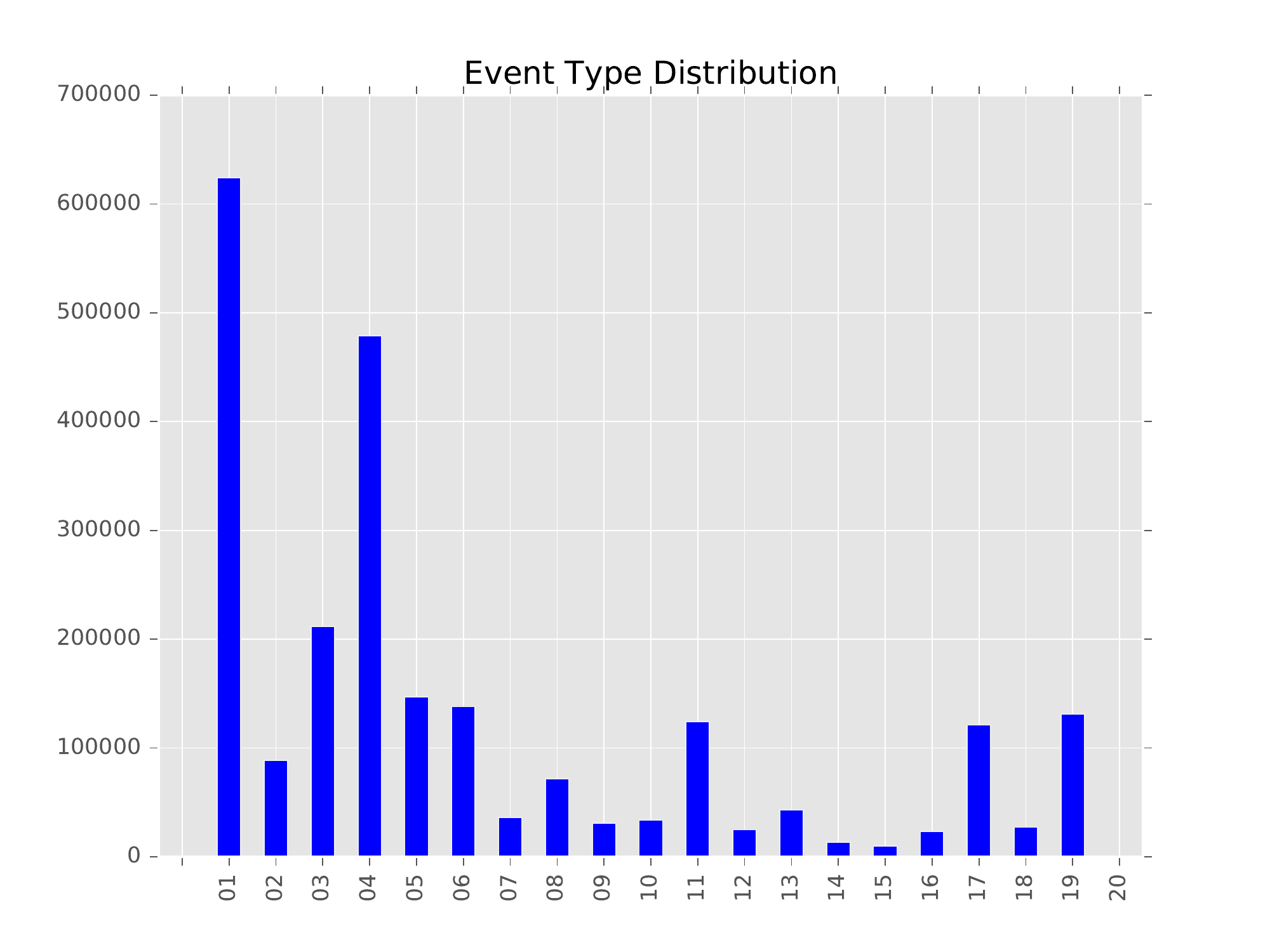}
\caption{Distribution of Event Types in Phoenix}
\end{figure}
\end{center}

Issue coding based on simple keyword lookups is used in Phoenix to provide further context to events. Figure 5 shows that the most 
common theme in the issue codings is terrorist organizations, followed by general security topics and the European Union. The hope
for these issue codings is that events that might not have clear actors can be further illuminated by an issue coding, such as
in the case of an attack against an unspecified armed group that could also have the issue coding of ``Terror Group.''

\begin{center}
\begin{figure}[H]
\includegraphics[width=6in]{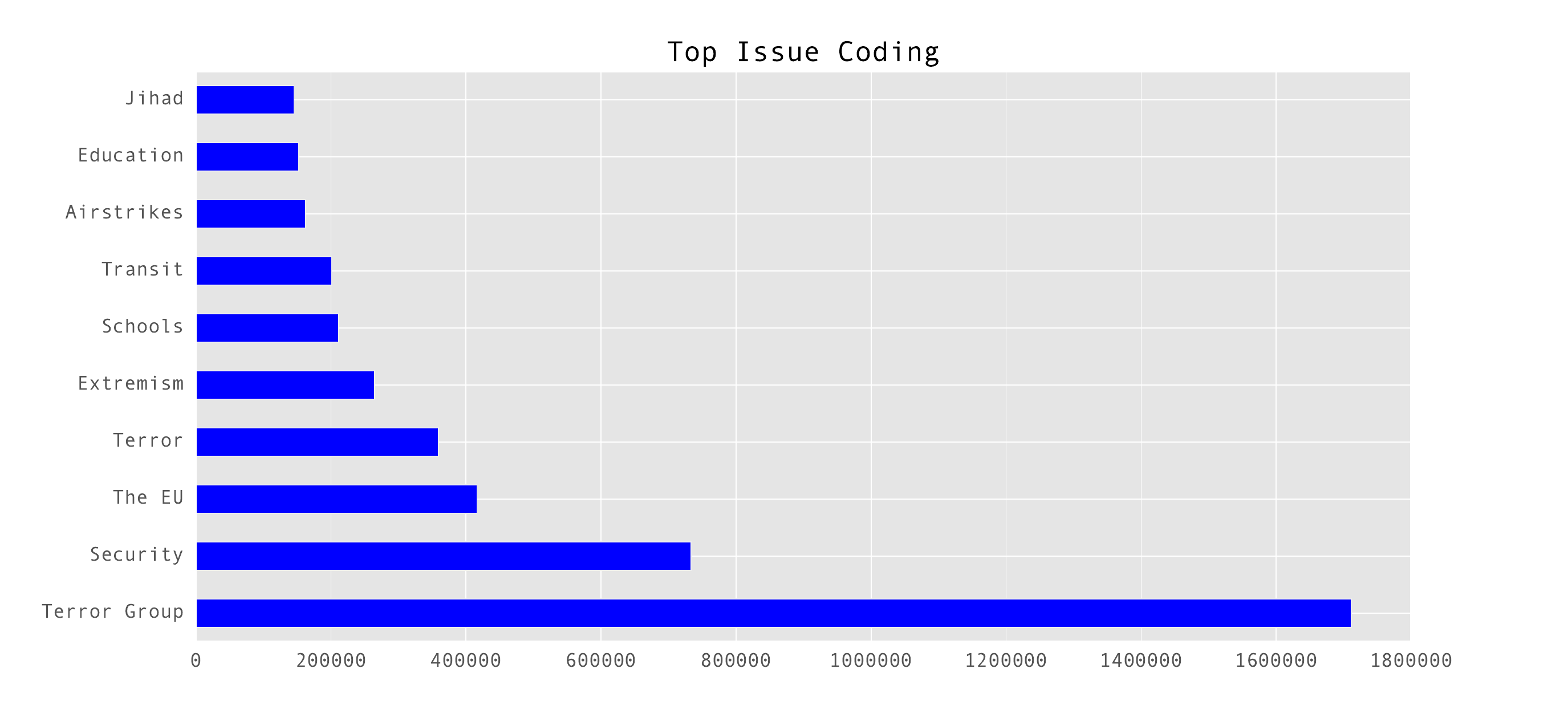}
\caption{Top 10 Issue Codings in Phoenix}
\end{figure}
\end{center}

\subsection{Actors}

Along the actor dimension, Figure 6 shows which full actor codings appear most often in the dataset. As one would expect, state
actors account for most of the events, with the only outlier the \texttt{IMGMOSISI} which is the actor code for the Islamic State in
Iraq and the Levant. This pattern also holds for just the entity codings, which could be either a state code or a few other important
codings such as IGOs. 

\begin{center}
\begin{figure}[H]
\includegraphics[width=6in]{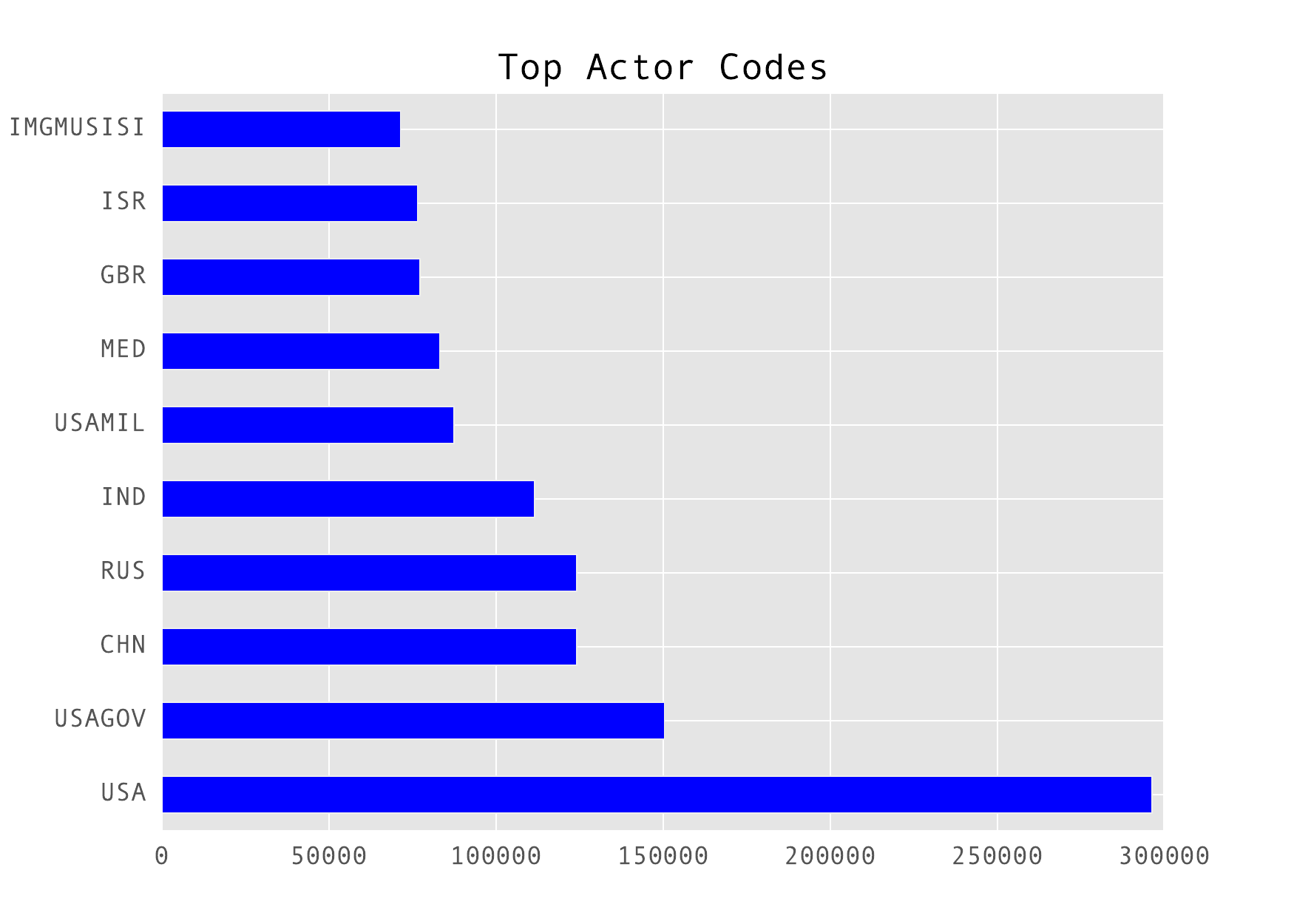}
\caption{Top 10 Full Actor Codings in Phoenix}
\end{figure}
\end{center}

\begin{center}
\begin{figure}[H]
\includegraphics[width=6in]{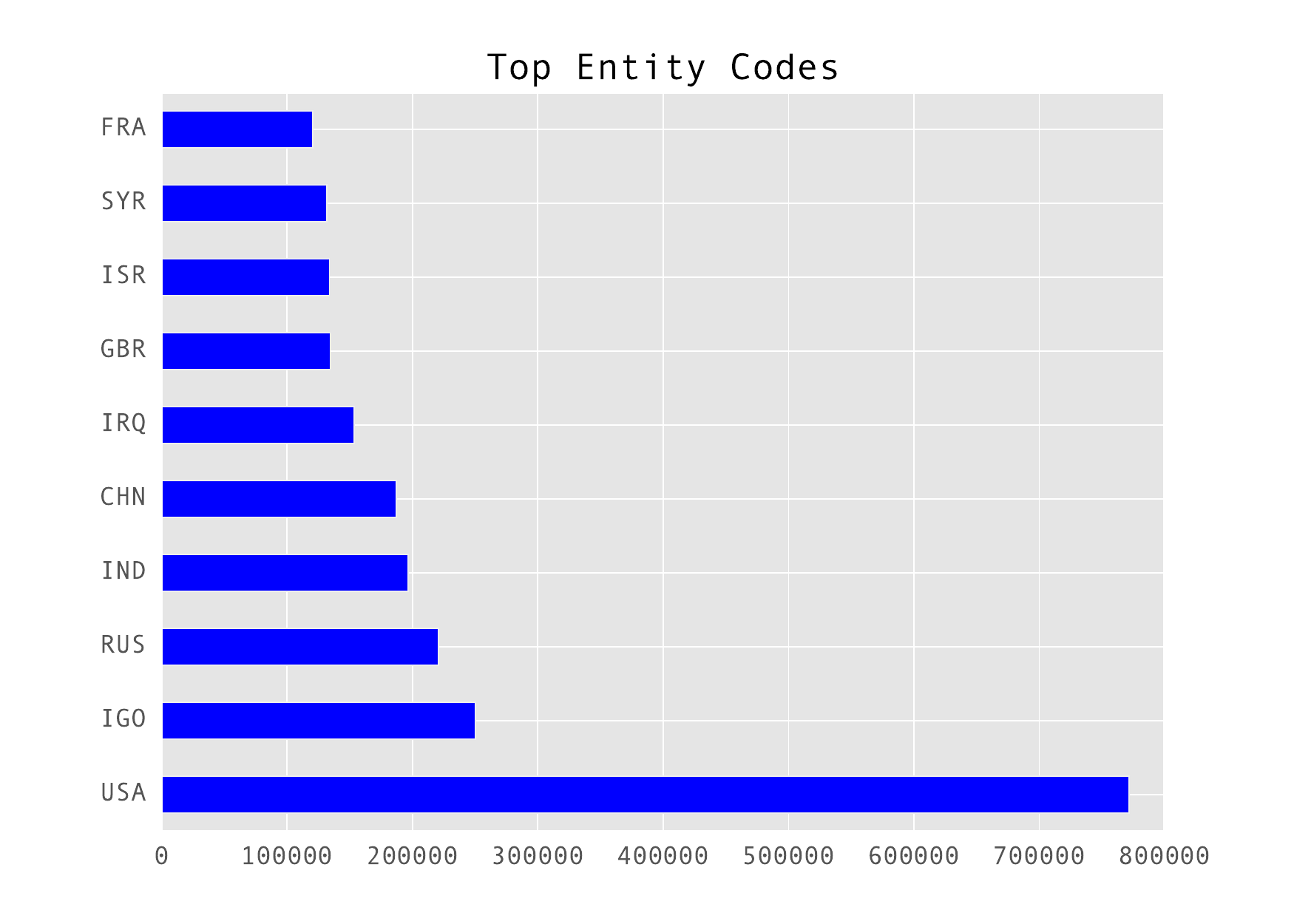}
\caption{Top 10 Entity Codings in Phoenix}
\end{figure}
\end{center}

It is possible to break the actor codes down further to examine role codes, which account for more specific functions that a specific
actor performs within a given country such as military or business. Figure 8 shows that the most common role code is government 
actors (\texttt{GOV}). Following the \texttt{GOV} role are military (\texttt{MIL}) and rebel (\texttt{REB}) codes. 

\begin{center}
\begin{figure}[H]
\includegraphics[width=6in]{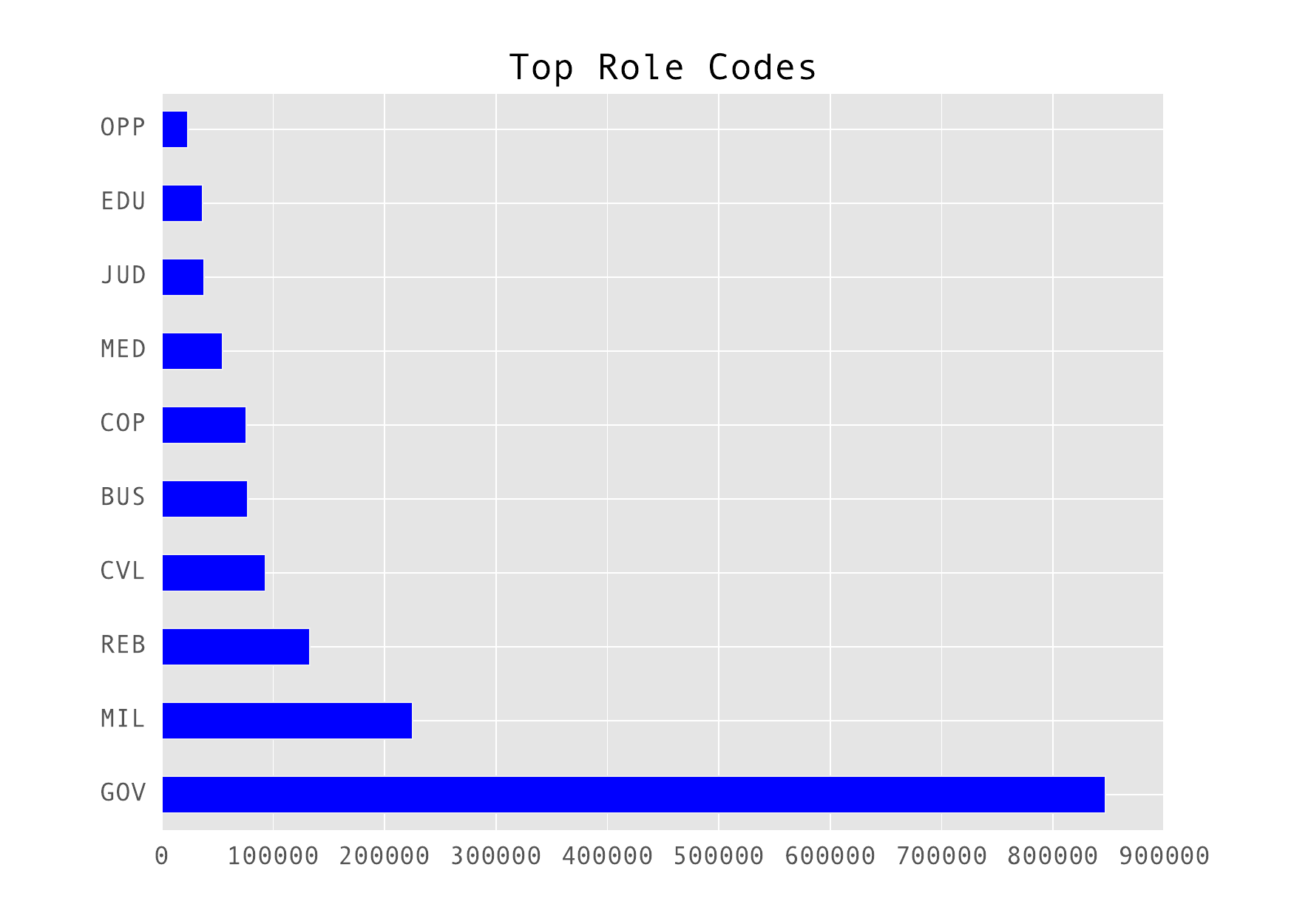}
\caption{Top 10 Role Codings in Phoenix}
\end{figure}
\end{center}

\subsubsection{Syria}

In order to better understand how the dataset is performing it is helpful to pull out a specific case and examine a similar set
of attributes as seen in the previous section. One of the major, ongoing events in the international arena during the time currently
covered by the Phoenix dataset is the conflict in Syria. Given this, I extract any events that contain the Syria country code, 
\texttt{SYR}, as the \texttt{SourceActorEntity} or \texttt{TargetActorEntity}. Figure 9 shows the plot of the daily aggregated event
counts. In this plot it is possible to see actions such as the beginning of United State intervention against ISIL, along with
other significant events within the country. As with any event data, it is important to note that the event counts shown do not
represent the on-the-ground truth of events in Syria, but instead reflect the media coverage of said events. Thus, some of the
peaks and troughs are the result of media coverage instead of any actual shift in reality. In order to provide more context to
the time series, Figure 10 shows the breakout of the \texttt{QuadClass} variable for this data subset. The dominant event types
are the low-level events described in the previous section, but the ``Material Conflict'' class is higher than in the broader
dataset. This is, of course, as expected given the ongoing conflict within Syria. 

\begin{center}
\begin{figure}[H]
\includegraphics[width=6in]{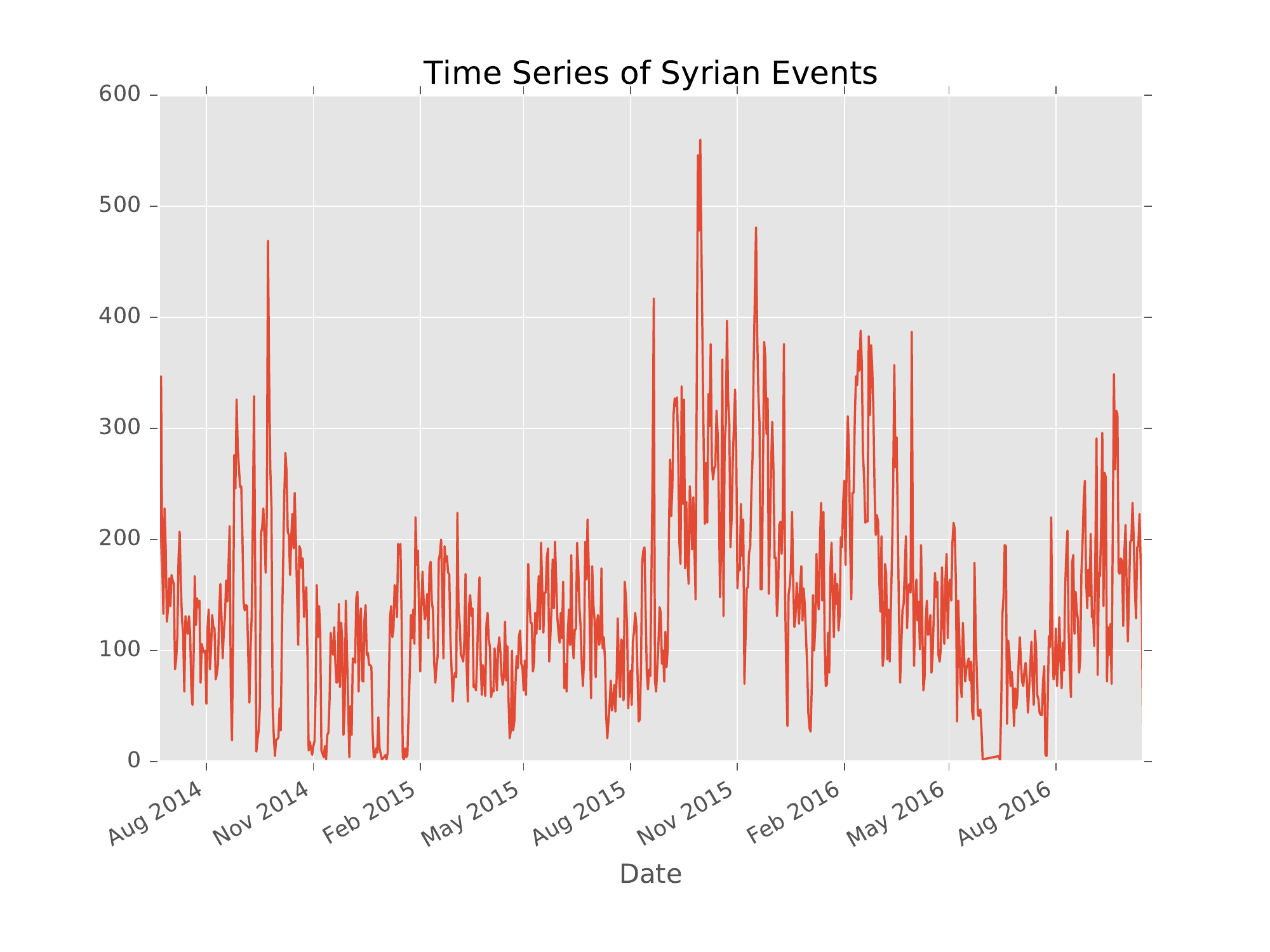}
\caption{Total Syrian Events In Phoenix Over Time}
\end{figure}
\end{center}

\begin{center}
\begin{figure}[H]
\includegraphics[width=6in]{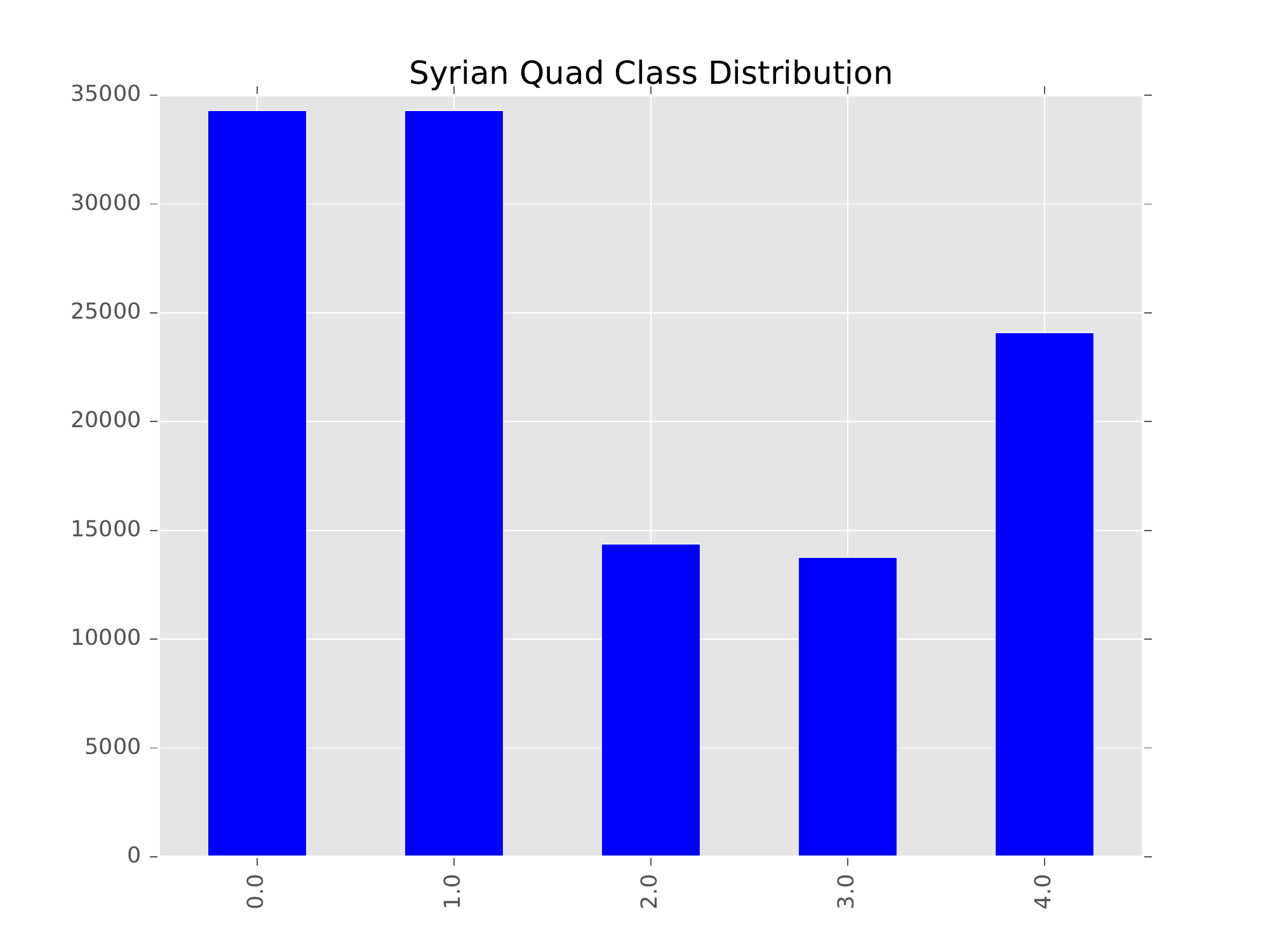}
\caption{Distribution of Quad Class Values for Syrian Events in Phoenix}
\end{figure}
\end{center}

Figures 11 and 12 show the top actor and entity codes for the Syria subset. Various Syrian actors appear most often, with
other Middle East countries also accounting for a fairly high portion of events. Also seen within this group of top actors is
ISIL and the United States. Additionally, Russia appears high in the
rankings of actors within Syria, capturing the recent activity by
Russian forces in support of the Assad regime.

\begin{center}
\begin{figure}[H]
\includegraphics[width=6in]{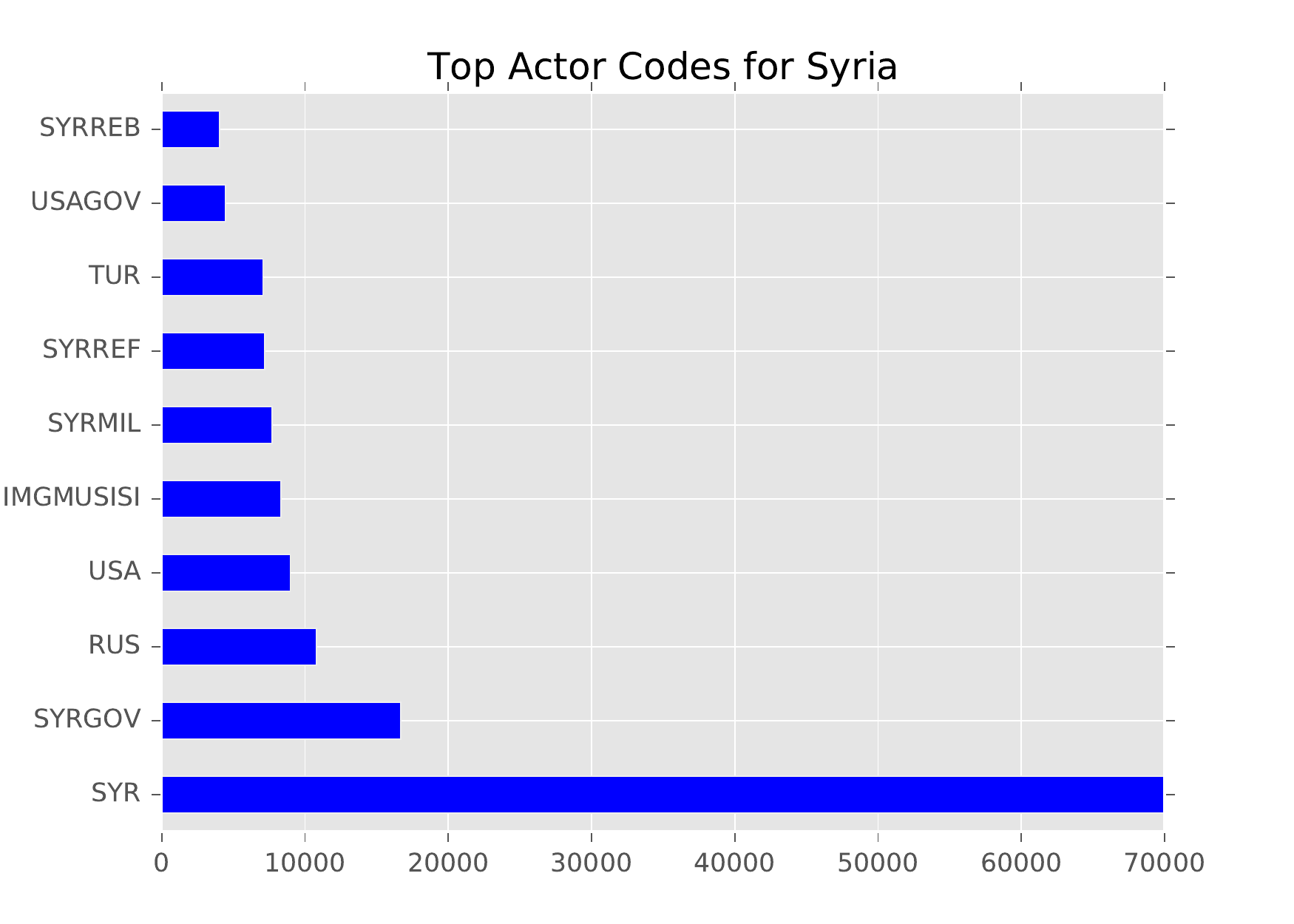}
\caption{Top 10 Full Actor Codings for Syrian Events in Phoenix}
\end{figure}
\end{center}

\begin{center}
\begin{figure}[H]
\includegraphics[width=6in]{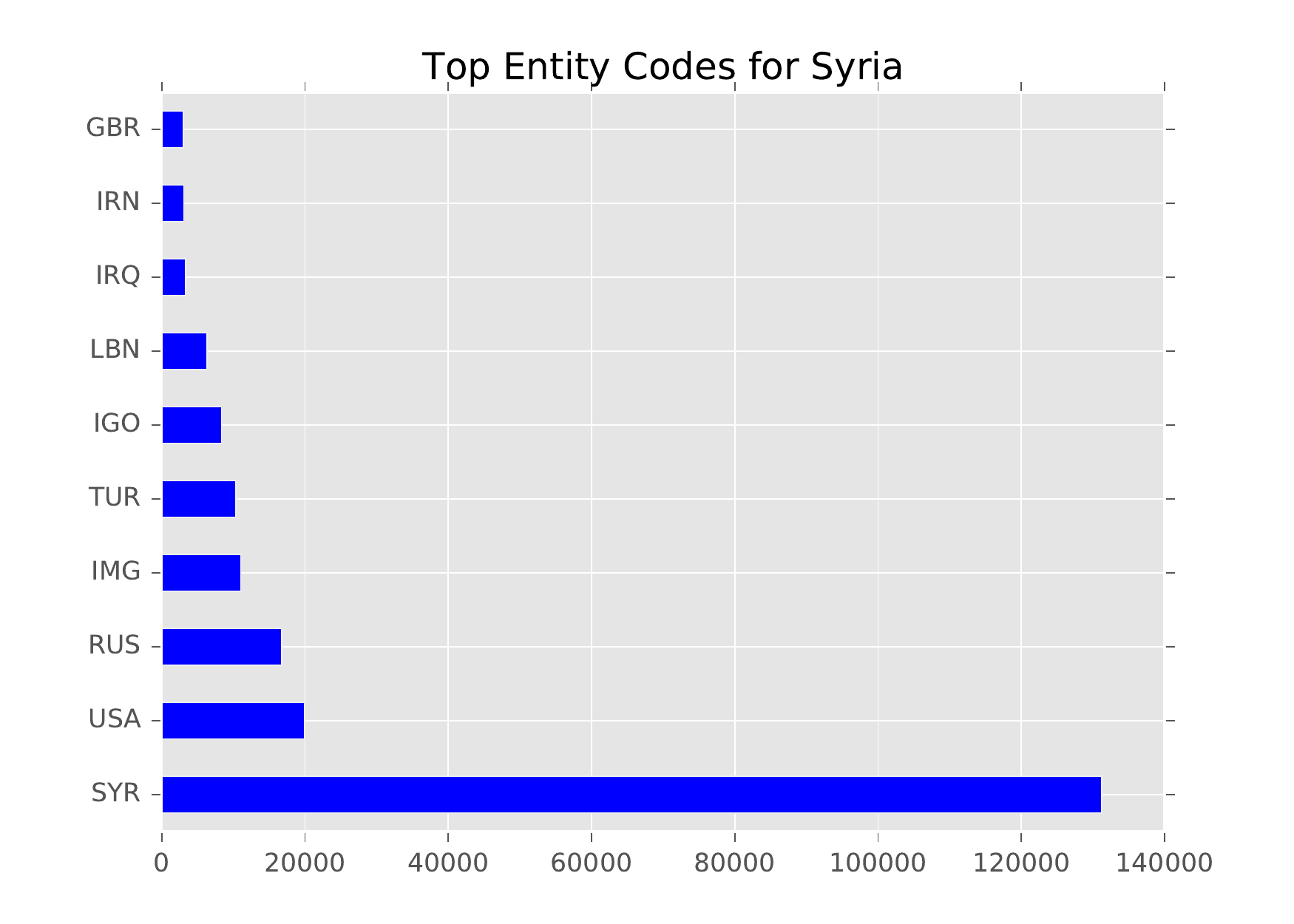}
\caption{Top 10 Entity Codings for Syrian Events in Phoenix}
\end{figure}
\end{center}


Overall, the Syrian subset of the data suggests that the Phoenix dataset is doing an adequate job of picking up events of
interest in a conflict zone. The actor and event distributions follow what one would expect for the Syrian conflict. Additionally,
there are no obvious errors in which actors make up the top participants in the subset. This examination provides confidence that the
dataset is ``working'' in terms of face validity. 

\subsection{Phoenix vs. ICEWS}

This section provides a comparison between the Phoenix dataset and the Integrated Crisis Early Warning System event dataset. The
comparison is at both the system and data level. That is, the following sections outline the differences and similarities in the
way ICEWS and Phoenix produce data, and how the generated data compares. The Phoenix data, as noted above, spans from June 2014 until
present day. ICEWS reaches further back into the past, with data starting in 1995, but the public data is subject to a one-year embargo.
This means that at the time of this writing (Fall 2016) there is roughly a year and a half of overlap between the two datasets. Thus,
the plots below show comparisons only during this time period. A final note relates to the existence, or lack thereof, of "gold standard"
records against which to compare the two datasets. \cite{growingPains} addresses this issue through the use of records coded
by the IARPA Open Source Indicators (OSI) program to serve as ground truth against which to compare ICEWS and GDELT.
These ground-truth observations are not publicly available at the current moment, though, so performing such a comparison
for Phoenix is beyond the reach of this dissertation.

\subsubsection{System}

The ICEWS project is similar in overall structure to the Phoenix data project: a real-time stream of news stories is ingested and processed
in various ways to create a final dataset of events. The stream of news stories ICEWS uses is made up of \citep{icewsCoding}:

\begin{quote}
[C]ommercially-available news sources from roughly 300 different publishers, including a mix of internationally (e.g., Reuters, BBC) and nationally (e.g., O Globo, Fars News Agency) focused publishers. The W-ICEWS program filters the data stream to those news stories more likely to focus on socio-political topics and less likely to focus on sports or entertainment.
\end{quote}

Additionally, the ICEWS project makes use of the BBN ACCENT coder. Since ACCENT is a propriety software produce developed by BBN, not much currently 
exists in the way of public description on how the coder works from an algorithmic perspective. Previous work by BBN on the SERIF coder does have
a public description, however, and it is likely that ACCENT shares something with the SERIF coder. \cite{serif} notes that SERIF works at both the sentence- and document-level to code events. At a high level, the coder makes use of a syntactic parse, and other linguistic information, to generate
text graphs with candidate who-did-what-to-whom relationships. The sentence-level information is aggregated up to a document-level in an attempt to
provide the most accurate event codings. The next section provides a comparison between the type of data the ICEWS coding procedure produces, and
the data the Phoenix pipeline produces.

\subsubsection{Data}

Figure \ref{ref:icews_phox} shows the plot of daily total events generated by Phoenix and ICEWS between June 2014 and late 2015.
Overall, the two datasets generate a remarkably similar number of events given the differing source materials and coding approaches as noted
in the previous section.
ICEWS shows more stability over time than Phoenix, with Phoenix not becoming fully stable until 2015. This is due to the ``beta'' nature
of much of the software underlying Phoenix until more focused developer support was available in 2015. The overall correlation
between the two series is .31, though this number is likely affected by the large swings in the Phoenix dataset. If days with less
than 1,000 events are dropped the correlation moves up to .49.

\begin{center}
\begin{figure}[H]
\includegraphics[width=6in]{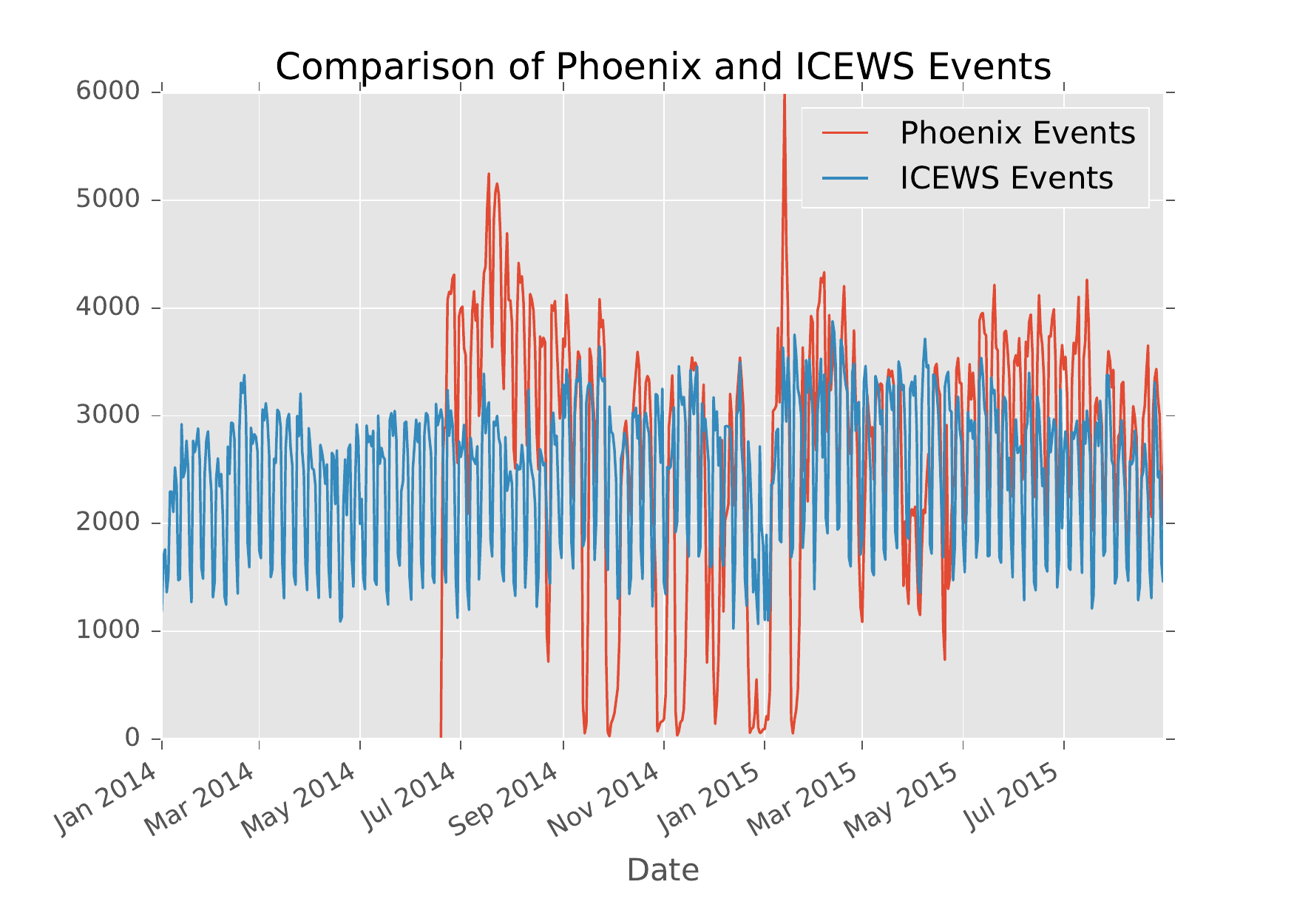}
\caption{Comparison of ICEWS vs. Phoenix Event Data}
\end{figure}\label{ref:icews_phox}
\end{center}

\begin{center}
\begin{figure}[H]
\includegraphics[width=6in, height=6in]{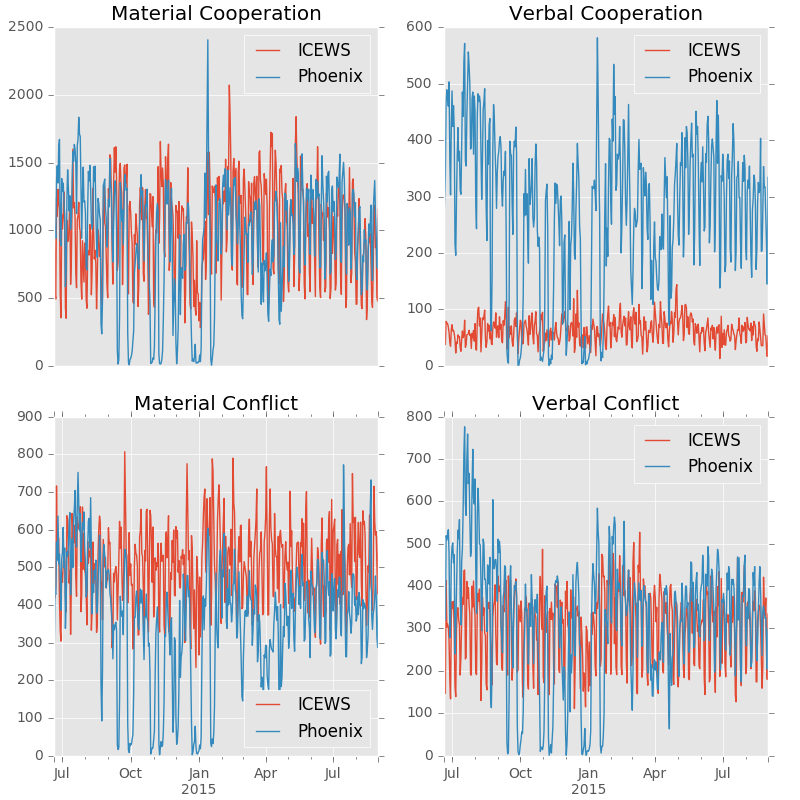}
\caption{Comparison of ICEWS vs. Phoenix Event Data}\label{ref:quad_series}
\end{figure}
\end{center}

Figure \ref{ref:quad_series} shows a pairwise comparison of each of the four QuadClass, excluding the ``Neutral'' category, as shown in
Table \ref{tab:cameo}. The main takeaway is that the broad trends appear
largely the same, though it is important to note the two categories that differ in a significant manner: ``Verbal Cooperation'' and ``Material
Conflict.'' These differences largely come down to implantation details that differ between the BBN ACCENT coder and the PETRARCH coder.\footnote{An additional difference can be seen in the way PETRARCH2 implements coding certain categories.} In short, the two coders implement slightly different
definitions of the various CAMEO categories based on a perception on the part of the designers or end-users as to what constitutes an interesting
and/or valid event within CAMEO. This point leads to a deeper discussion as to what, exactly, constitutes the CAMEO coding ontology; Chapter 5 contains a deeper discussion of these issues.

While it is not currently possible to make definitive judgements as to which dataset most closely captures ``truth'', another point more deeply
discussed in Chapter 5, it is interesting to note that the statistical signal contained within the two datasets, as evidenced by the correlations
and broad trends, is not largely different.

\section{Conclusion}

This paper has shown that creating a near-real-time event dataset, while using deep parsing methods and advanced natural language
processing software, is feasible and produces useful results. The combination of various technological and software advances enables
a new generation of political event data that is distinctly different from previous iterations. In addition to the advances in
accuracy and coverage, the marginal cost of generating event data is now nearly zero. Even with previous automated coding efforts,
human intervention was necessary to gather and format news content. With the addition of real-time web scraping, the entire system 
has moved much closer to a ``set it and forget it'' model. The primary interaction needed once the system is running is to 
periodically check to ensure that relevant content is scraped and that no subtle bugs cause the system to crash.

While this new generation provides an improvement over previous iterations, there is still much work to be done. The main 
place for future work is deeper integration with the open-source NLP software. The PETRARCH system currently uses the parse
information provided by CoreNLP to distinguish noun and verb phrases. This is actually a fraction of the information provided
by CoreNLP. Additional information includes named entity recognition and a semantic dependency parse, which shows how words
relate to each other in a more complex way than in the standard parse tree \citep{syntactic}. Using this information would allow for a more
accurate event coding since events could be constructed in a manner that fits better with the natural construction of a sentence.
Additionally, using a semantic dependency parse could alleviate issues of constructing arbitrary actor codings since codes
would be built based on noun-adjective relationships. When combined with named entity recognition this could prove to be a quite
powerful approach. 

\newpage
\bibliographystyle{apsr}
\bibliography{Biblio-Database}

\end{document}